\documentclass[a4paper,UKenglish,cleveref, autoref, thm-restate]{article}

\usepackage{algorithm2e}
\usepackage{amsmath}
\usepackage{amsthm}
\usepackage{amsfonts}
\usepackage{pgfplots}
\pgfplotsset{compat=1.18}
\usepackage{hyperref}
\usepackage{multicol}
\usepackage{multirow}

\usepackage{lscape}
\usepackage{rotating}

\newcommand{\U}{{\cal U}}
\newcommand{\V}{{\cal V}}

\newcommand{\C}{{\cal K}}
\newcommand{\GC}{{\cal GC}}

\newcommand{\K}{{Cores}}

\newtheorem{observation}{Observation}

\title{Empirical Evaluation of the Implicit Hitting Set Approach for Weighted CSPs}
\author{
Aleksandra Petrova  \\
\and Javier Larrosa \\
\and Emma Rollón \\
\multicolumn{1}{p{.8\textwidth}}{\centering\emph{Department of Computer Science, Universitat Politècnica de Catalunya, Spain}\\
}}

\begin{document}

\maketitle
\begin{abstract}
SAT technology has proven to be surprisingly effective in a large variety of domains. However, for the  \textit{Weighted} CSP problem dedicated algorithms have always been superior. 
One approach not well-studied so far is the use of SAT in conjunction with the \textit{Implicit Hitting Set} approach. In this work, we explore some alternatives to the existing algorithm of reference. The alternatives, mostly borrowed from related boolean frameworks, consider trade-offs for the two main components of the IHS approach: the computation of low-cost hitting vectors, and their transformation into high-cost cores. For each one, we propose 4 levels of intensity. Since we also test the usefulness of cost function merging,  our experiments consider 32 different implementations. Our empirical study shows that for WCSP it is not easy to identify the best alternative. Nevertheless, the cost-function merging encoding and extracting maximal cores seems to be a robust approach.
\end{abstract}

\section{Introduction}
\label{sect:introduction}

The \textit{Weighted CSP} problem (WCSP) is a framework for discrete optimization with many practical applications \cite{DBLP:journals/constraints/CabonGLSW99,DBLP:journals/bioinformatics/ViricelGSB18,DBLP:journals/bioinformatics/VucinicSRBS20,DBLP:journals/constraints/BensanaLV99,DBLP:journals/constraints/CabonGLSW99, DBLP:journals/bioinformatics/ViricelGSB18,DBLP:journals/bioinformatics/VucinicSRBS20,DBLP:journals/constraints/BensanaLV99} that has attracted the interest of researchers for decades  \cite{DBLP:journals/ai/CooperGSSZW10,DBLP:journals/ai/LarrosaS04,DBLP:conf/cp/AlloucheGKSZ15,DBLP:conf/cp/BeldjilaliMAKG22}.
In this paper, we focus on the \textit{Implicit Hitting Set Approach} (IHS) for WCSP solving. The idea of the
IHS algorithms is to iteratively grow a set of unsatisfiable pieces of the problem (called \textit{cores}) and find if it is possible to solve the problem by avoiding (\textit{i.e}, \textit{hitting}) them. The algorithm terminates when the best way to hit the identified cores incidentally also hits the unidentified cores.

The motivation for our work is that IHS is surprisingly effective for the MaxSAT problem \cite{DBLP:phd/ca/Davies14,DBLP:conf/cp/DaviesB13,DBLP:conf/sat/BergBP20}, and such success has been lifted to several generalization frameworks such as \textit{Pseudo-boolean} optimization \cite{DBLP:conf/cp/IHS-PB1,DBLP:conf/sat/IHS-PB2}, \textit{MaxSMT} \cite{DBLP:conf/cade/FazekasBB18} and \textit{Answering Set Programming} \cite{DBLP:conf/kr/SaikkoDAJ18}.
Although MaxSAT and WCSP are fairly similar, to the best of our knowledge the only works that consider IHS for WCSP are \cite{DBLP:conf/cp/DelisleB13,Larrosa24} and, in both cases, a very simple IHS strategy is used.

 In this paper, we test the potential of several variations of the original algorithm that have already been proposed and successfully applied in other frameworks. Here, we adapt them to WCSPs and test their effect empirically. 
 We consider four different ways in which to obtain hitting vectors and four different ways in which to transform them into an improved core. Since we also test the effectiveness of \textit{cost-function merging} \cite{Larrosa24} we end up with $4\times 4 \times 2 =32$ different algorithms that we test on several benchmarks.
 
 The experiments show that different algorithms may have a dramatic difference in performance (up to several orders of magnitude). Most importantly, they show that no algorithm systematically dominates all the others, but merging cost-functions following \cite{Larrosa24} and computing maximal cores seems to be the most robust strategy. This configuration most of the times is the most efficient or is close to the most efficient. Regarding the computation of hitting vectors, greedy heuristics often do not pay off. Although there is no clear winner between optimal and cost-bounded hitting vectors, cost-bounded ones seem to have more potential because it offers more implementation alternatives.
 

\section{Preliminaries}\label{sec:prelim}

A \textit{Constraint Satisfaction Problem} (CSP) is a pair $(X,C)$ where $X$ is a set of \textit{variables} taking values in a finite domain, and $C$ is a set of \textit{constraints}. Each constraint depends on a subset of variables called \textit{scope}. Constraints are boolean functions that forbid some of the possible assignments of the scope variables. A \textit{solution} is an assignment to every variable that satisfies all the constraints. Solving CSPs is known to be an NP-complete problem \cite{DBLP:reference/fai/GentPP06}.

A \emph{Weighted CSP} (WCSP) is a CSP augmented with a set $F$ of \emph{cost functions}.  A cost function $f\in F$ is a mapping that associates a cost to each possible assignment of the variables in its scope. The \emph{cost of a solution} is the sum of costs given by the different cost functions. The WCSP problem consists of computing a \textit{solution of minimum cost} $w^*$. 

The IHS approach for WCSP is defined in terms of vectors. The \textit{cost} of vector $\vec{v}=(v_1,v_2,\ldots,v_m)$ is $cost(\vec{v})=\sum_{i=1}^m v_i$.
In
the (partial) order among same-size vectors, $\vec{u}\leq \vec{v}$, holds iff for each component $i$ we have that $u_i\leq v_i$.  If $\vec{u}\leq \vec{v}$ we say that
 $\vec{v}$ \textit{dominates} $\vec{u}$. We say that a set of vectors $\V$ \textit{dominates} a vector $\vec{u}$, noted $\vec{u}\leq \V$, if there is some $\vec{v}\in \V$ that dominates $\vec{u}$. Further, we say that a set of vectors $\V$ \textit{ dominates} a set of vectors $\U$, noted $\U\leq \V$, if $\V$ dominates every vector of $\U$.
Given a set of vectors $\U$, a vector $\vec{u}\in \U$ is \textit{maximal} if it is not dominated by any other element of $\U$. The set of maximal vectors in $\U$ is noted $\overline{\U}$. The set of vectors in $\U$ with cost less than $w$ is noted $\U_{(<w)}$. 
If $\vec{u}$ is \textit{not} dominated by $\V$ we say that it \textit{hits} $\V$.  
The \textit{minimum cost hitting vector} MHV of $\V$ is a vector that hits $\V$ with minimum cost. It is not difficult to see that MHV reduces to the classic \textit{minimum hitting set problem} \cite{DBLP:books/fm/GareyJ79}, which is known to be NP-hard.

In the following, we will consider an arbitrary WCSP $(X,C,F)$ with $m$ cost functions $F=\{f_1,f_2,\ldots,f_m\}$. 
A \textit{cost vector} $\vec{v}=(v_1,v_2,\ldots,v_m)$ is a vector where each component $v_i$ is associated to cost function $f_i$, and value $v_i$ must be a cost occurring in $f_i$. Cost vector
$\vec{v}$ \textit{induces} a CSP $(X,C \cup F_{\vec{v}})$ where $F_{\vec{v}}$ denotes the set of constraints $(f_i\leq v_i)$ for $1\leq i \leq m$ (namely, cost functions are replaced by constraints). If the  CSP induced by $\vec{v}$ is satisfiable we will say that $\vec{v}$ is a \textit{solution vector}. Otherwise, we will say that $\vec{v}$ is a \textit{core}. The set of all cores will be denoted $\K$. 
An \textit{optimal solution vector} is a solution vector of minimum cost. 
It is easy to see that the cost of an optimal solution vector is the same as the optimum cost $w^*$ of the WCSP. 

\section{IHS-based WCSP solving}\label{sec:ihs}

The IHS approach relies on the following observation that establish a lower bound and an upper bound condition in terms of cores and solutions,
\begin{observation}\label{obs1}
Consider a solution vector $\vec{h}$ and a set of cores $\C \subseteq \K$. Then, $MHV(\C)\leq w^* \leq cost(\vec{h})$.
\end{observation}

All the algorithms discussed in this paper will aim at finding a solution $\vec{h}$  and a (possibly small) set of cores $\C$ such that the two bounds meet (that is, $MHV(\C)=cost(\vec{h})$). 
This condition corresponds to $\vec{h}$ being optimal and $\C$ being the proof of its optimality. We will refer to this (termination) condition as TC. 

Consider the set of all cores with costs less than $w^*$.
We define a \textit{goal core} as a maximal core in that set. That is, the set of goal cores is $\GC= \overline{\K}_{(<w^*)}$
The following observation rephrases the lower bound part of TC as having a set of cores $\C$ that dominates all maximal goal cores,

 \begin{observation}\label{obs2}
 A set of cores $\C$  satisfies  $MHV(\C)=w^*$ if and only if  $\GC \leq \C$, where $\GC$ denotes the set of goal cores. 
 \end{observation}

 Therefore, IHS algorithms must compute a set $\C$ that dominates every goal core (lower bound condition of TC) and an optimal solution (upper bound part of TC).

\subsection{Baseline Algorithm}\label{subsec:ihslb}

\begin{algorithm}[t]
  \LinesNumbered

  \SetKwFunction{SolveWCSP}{IHS-$lb$}
  \SetKwFunction{Solve}{SolveCSP}
  \SetKwFunction{SolveMin}{MinCostHV}
  \SetKwFunction{Grow}{ImprCore}
\begin{multicols}{2}
  
  {\bf Function} \SolveWCSP{$X,C,F$}\\
  \Begin{
    $\C:=\emptyset;\ lb:=0;\ ub:=\infty\ $\;
    \While{$lb<ub$}{
$\vec{h}:=$\SolveMin{$\C$}\;
$lb:=cost(\vec{h})$\;
\lIf{\Solve{$X,C \cup F_{\vec{h}}$}}{$ub:=cost(\vec{h})$}
    \Else{
     $\vec{k}:=$\Grow{$X,C,F,ub,\vec{h}$}\;
     $\C:=\C \cup \{\vec{k}\}$\;     
    }
}
\Return $lb$
}
\columnbreak
  \SetKwFunction{SolveWCSP}{IHS-$ub$}
  \SetKwFunction{Solve}{SolveCSP}
  \SetKwFunction{SolveMin}{CostBoundedHV}
  \SetKwFunction{Grow}{ImprCore}

  {\bf Function} \SolveWCSP{$X,C,F$}\\
  \Begin{
    $\C:=\emptyset;\  lb:=0;\ ub:=\infty\ $\;
    \While{$lb<ub$}{
$\vec{h}:=$\SolveMin{$\C,ub$}\;
\lIf{$\vec{h}=NUL$}{$lb:=ub$}
\Else{
\lIf{\Solve{$X,C \cup F_{\vec{h}}$}}{$ub:=cost(\vec{h})$}
\Else{
     $\vec{k}:=$\Grow{$X,C,F,ub,\vec{h}$}\;
     $\C:=\C \cup \{\vec{k}\}$\;     
    }
    }
}
\Return $lb$
}
\end{multicols}
\caption{Two different IHS algorithms for WCSP. Both receive as input a WCSP $(X,C,F)$ and returns the cost of the optimal solution $w^*$. Function \textit{ImprCore()} receives as input a core $\vec{h}$ and returns a core $\vec{k}$ such that $\vec{h}\leq \vec{k}$. It may also improve the upper bound $ub$.}
  \label{alg:ihs1}
\end{algorithm}

The algorithm proposed in \cite{DBLP:conf/cp/DelisleB13} appears at the left of 
Algorithm \ref{alg:ihs1}. It is a loop that maintains three variables: a working set of cores $\C$, a lower bound $lb$, and an upper bound $ub$ of the optimum. At each iteration, the algorithm computes in $\vec{h}$ the MHV of $\C$ and solves the CSP that it induces. If it is satisfiable the algorithm will stop, else $\vec{h}$ is improved into core $\vec{k}$, which is added to $\C$, and the algorithm goes on. 
The details for \textit{ImprCore()} will be discussed in Subsection \ref{subsec:cores}. For the moment, just note that
it returns a core vector $\vec{k}$ such that $\vec{h}\leq \vec{k}$.  \textit{ImprCore()}  may find solution vectors during its execution. If their cost is smaller than the 
upper bound, the upper bound will be accordingly updated. Because the emphasis of this algorithm is in the lower bound (the upper bound is only updated if better solutions are found incidentally) we will refer to it as IHS-$lb$.

It is worth noting at this point that each iteration of IHS can be divided into two parts: $i$) the computation of the hitting vector $\vec{h}$ which requires solving an NP-hard optimization problem and $ii$) if $\vec{k}$ is a core, its transformation into a larger core $\vec{k}$ which requires solving a sequence of CSPs, which are NP-complete decision problems. In this paper, we restrict ourselves to the usual choice of computing hitting vectors with a 0/1 IP solver and solving induced CSPs with a SAT solver.

\section{Algorithmic Alternatives}

\subsection{Computation of Hitting Vectors}\label{subsec:hv}
In the following, we describe some alternatives to alleviate the time spend computing hitting vectors.

\subsubsection{Non-optimal Cost-bounded Hitting Vectors}

 One way to decrease the workload of each iteration is to rely on non-optimal hitting vectors. As suggested in \cite{DBLP:conf/sat/IHS-PB2}, we can replace optimal hitting vectors by hitting vectors of bounded cost. The right side of Algorithm \ref{alg:ihs1} shows this idea. At each iteration, a hitting vector $\vec{h}$ with cost less than $ub$ is obtained.  If $\vec{h}$ is a solution the upper bound is updated, else it is improved and added to $\C$. Since the emphasis of this algorithm is in the upper bound, we will refer to it as IHS-$ub$.

The main advantage of IHS-$ub$ compared to IHS-$lb$ is that iterations are likely to be faster. There are several reasons for this. On the one hand, it is much more efficient to find a bounded hitting vector which is a decision problem, than finding an optimal hitting vector which is an optimization problem. On the other hand, only in the last call of \textit{CostBoundedHV()} the problem will be unsatisfiable which is typically a much more costly task to solve. Another reason is that cost-bounded hitting vectors are likely to have a higher cost (near $ub$) and therefore obtaining an improved core $\vec{k}$  will not need so many \textit{SolveCSP()} calls (whatever the stopping criteria are).
However, the advantage is at the cost of potentially more iterations. On the one hand in IHS-$ub$, not all iterations end up adding a new core because some iterations only decrease the upper bound. On the other hand, the computed hitting vectors are no longer guaranteed to have cost below $w^*$, therefore new cores may not contribute towards the lower bound part of TC (that is, $\GC \leq \C$).

\subsubsection{Cost-unbounded Hitting Vectors}

Although obtaining cost-bounded hitting vectors can be done much more efficiently in practice than obtaining optimal hitting vectors, the problem remains NP-complete and, therefore, may still be time-consuming. As suggested in \cite{DBLP:phd/ca/Davies14} and subsequently applied in \cite{DBLP:conf/cp/IHS-PB1}, one way to avoid expensive calls is by removing the cost-bound requirement. Obtaining a low-cost hitting vector without requiring the cost to be below a bound can easily be done with an incomplete algorithm. 

We consider a greedy algorithm that starts from $\vec{h}=\vec{0}$ and makes a sequence of increments dictated by some greedy criterion until $\vec{h}$ hits every vector in $\C$.  Since we want to hit as many cores as possible with the lowest cost, our algorithm selects the increment that minimizes the corresponding ratio. This algorithm is similar to a well-studied greedy algorithm for \textit{vertex covering} \cite{cormen01introduction}.

If the resulting hitting vector $\vec{h}$ is a core, then the algorithm can do the usual core improvement adding $\vec{k}$ to $\C$ as it would have happened with either IHS-$lb$ or IHS-$ub$. If $\vec{h}$ is a solution there are two cases. If its cost is less than $ub$, the upper bound is updated as it would have happened with IHS-$ub$. Alternatively, if its cost is more than or equal to $ub$, then there is no use for $\vec{k}$ and the iteration has been useless. To avoid the algorithm entering infinite loops, \cite{DBLP:phd/ca/Davies14} suggests forcing the following iteration not to use the greedy algorithm. In our case, depending on whether the next iteration computes an optimal hitting vector or a cost-bounded hitting vector we will denote the algorithm IHS-$grdlb$ or IHS-$grdub$.

\subsection{Improving Cores}\label{subsec:cores}

The different alternatives considered so far aimed at alleviating the time spent in computing hitting vectors. Now we address the task of improving cores. Adding to $\C$ cores with high values in their components is beneficial because they 
will increase the set of cores that they dominate and they will likely contribute towards the lower bound part of the Termination Condition. However, computing high-cost cores is more time consuming, and the right trade-off must be found.

We are restricting ourselves to algorithms that depart from a hitting vector $\vec{h}$ and make a sequence of greedy increments to its components until some stopping criteria are achieved while preserving the core condition. In the following, we consider four alternatives.

\begin{itemize}
    \item \textit{Maximal Cores.}
The strongest (and most time consuming) criterion is \textit{core maximality} (that is, achieving a vector that cannot be increased in any of its components without losing the core condition). Our algorithm is reminiscent of the so-called \textit{destructive MUS extraction} \cite{DBLP:conf/sat/MUSalg}. It uses a set $I$ that contains the list of indices that may be increased. While the set $I$ is not empty, an index is selected and its component is increased. If the resulting vector is not a core, the increment is undone and the indexed is removed from $I$. If a component cannot be further increased because it has reached its maximum cost, the indexed is also removed from $I$.

\item \textit{Lazy Cores.} The weakest (and least time consuming) option for improving cores is not to do any improvement of $\vec{h}$ whatsoever. However, since our implementation of \textit{SolveCSP()} uses an assumptions-based SAT solver, we can take advantage of the core improvement that it gives us for free as part of the resolution.

\item \textit{Cost-bounded Cores.} One intermediate option is to improve cores until their cost reaches the upper bound. If the cost of $\vec{h}$ is less than the optimal value $w^*$ and the cost of the resulting core $\vec{k}$ is more than or equal to $w^*$, then we know that adding $\vec{k}$ to the set of cores $\C$ will contribute towards the lower bound part of the TC because at least one more goal core will be dominated. Obviously, during the execution, we do not know the value of $w^*$, but hitting vectors computed by any of the four algorithms, at least at early iterations, are likely to have cost less than $w^*$. Then, if we improve them until their cost equals $ub$, the new set of cores will certainly have contributed towards TC. 

\item \textit{Partially Maximal Cores.} Another alternative between lazy and maximal cores that was already used in \cite{DBLP:conf/cp/DelisleB13} is to soften the condition of core maximality and request that the increment of just one component (instead of all of them) produces a solution vector.

\end{itemize}

In all our implementations of \textit{ImprCore()}, we select at each iteration the index $i$ with the lowest value $v_i$ among the set of candidates. The reason is that having in $\C$ cores whose minimum value is as large as possible makes it more costly to be hit, which, in turn, makes the lower bound grow faster.

\section{Empirical Results}\label{sec:exp}

\begin{table}[h]
\begin{center}
\begin{tabular}{ |l|c|c|c|c|c| } 
 \hline
 \textbf{Problem} & \textbf{Variables} & \textbf{Max dom size} & \textbf{Constraints} & \textbf{Cost functions} & \textbf{Max costs}\\
 \hline
 Rnd domains & 16 - 23 & 30 & 44 - 59 & 44 - 59 & 6 - 9\\
 \hline
 Rnd weights & 16 - 23 & 5 & 48 - 68 & 48 - 68 & 20 - 21\\
 \hline
 Rnd sparse & 34 - 46 & 5 & 99 - 122 & 99 - 122 & 7 - 9\\
 \hline
 Rnd scale-free-4 & 23 - 25 & 5 & 92 - 99 & 92 - 99 & 7 -9\\
  \hline
 Rnd scale-free-5 & 24 - 25 & 5 & 107 - 114 & 107 - 114 & 8 -9\\
  \hline
 Ehi & 297 - 315 & 7 & 4081 - 4400 & 4081 - 4400 & 2\\
 \hline
 SPOT5 & 45 - 506 & 4 & 122 - 9325 & 41 - 440 & 2 - 3\\
 \hline
 driverlog &  46 - 546 & 4 - 12 & 156 - 14429 & 55 - 702 & 3 - 7\\
 \hline
 Grid & 396 - 400 & 2 & 757 - 801 & 757 - 801 & 3\\ 
 \hline
 Normalized & 101 - 8621 & 2 & 227 - 19903 & 57 - 195 & 2 - 10\\ 
 \hline
 Pedigree & 208 - 9403 & 3 - 10 & 447 - 33795 & 180 - 10621 & 3 - 17\\ 
 \hline
 CELAR &  13 - 222 & 14 - 44 & 65 - 944 & 65 - 876 & 11 - 137\\ 
 \hline
\end{tabular}
\end{center}
    \caption{Summary of the benchmarks main characteristics: range of variables, largest domains, number of constraints, number of cost functions and maximum number of different costs appearing in the cost functions. Note that these values have been collected after VAC pre-processing.}
    \label{table:benchmark}
\end{table}

The experiments reported ran on nodes with 4 cores 16Gb \textit{Dell} \textit{PowerEdge} R240 with \textit{Intel Xeon} E-2124 of 3.3Ghz. \textit{MinCostHV()} and \textit{CostBoundedHV()} were modeled as 0-1 integer programs and solved with \textit{CPLEX} \cite{cplex2009v12}. Induced CSPs were encoded as CNF SAT formulas and solved with \textit{CaDiCaL} \cite{cadical}.

For the experiments, we used several benchmarks aiming at a heterogeneous sample of instances. Table \ref{table:benchmark} summarizes the features of each group of instances. 
All instances are pre-processed and made \textit{virtually arc consistent} (VAC)~\cite{DBLP:journals/ai/CooperGSSZW10}. Unless indicated otherwise, we use the \textit{cost-function merging} formulation proposed in~\cite{Larrosa24} where clusters of cost functions are heuristically determined using a tree decomposition and (virtually) merged into a single cost function. Also, our implementation may compute several \textit{disjunctive cores} in the same iteration as proposed in~\cite{DBLP:conf/cp/DelisleB13}. All executions had a time out of one hour. 
We conducted the empirical evaluation over the following benchmarks:

\textit{Uniform random} instances are characterized by five parameters $(n,d,m,w,t)$ that correspond to the number of variables, domain size, number of binary cost functions, number of different weights at each cost function and number of tuples with non-zero cost at each cost function, respectively. The scope of the $m$ (out of $\frac{n(n-1)}{2}$ alternatives) cost functions, the $t$ (out of $d^2$ alternatives) tuples with non-zero cost and their actual cost (out of the $w$ alternatives)  are decided using a uniform random distribution. 
We generated 3 groups of instances aiming at increasing one of the parameters: \textit{domains} $(25,30,50,5,750)$,  \textit{weights} $(25,5,50,10000,20)$, and \textit{sparse} $(50,5,100,5,20)$. For each group we generated $50$ instances. 

Uniform random instances have been long used for testing purposes, but they are sometimes questioned because real instances are anything but random. A more realistic graph structure that has been used for empirical testing are \textit{scale-free graphs} \cite{DBLP:journals/algorithms/AnsoteguiBL22}. It is known that scale-free networks appear in many real-world networks like the World Wide Web, some social networks like papers co-authorship or citation, protein interaction network, \textit{etc}. In our scale-free instances the constraint graph is a scale-free graph following the \textit{Barabási-Albert} model. Instances are also characterized by five parameters $(n,d,m,w,t)$ but unlike uniform random instances, $m$ refers to the model's parameter. We report results for the classes $(25,5,4,5,20)$ and $(25,5,5,5,20)$.

Finally, we selected miscellaneous instances from the well-known \textit{evalgm} repository\footnote{http://genoweb.toulouse.inra.fr/~degivry/evalgm/} that were within the reach of IHS-based algorithms. The selection includes includes: 
\textit{EHI} (Random 3-SAT instances embedding a small unsatisfiable part and coverted into a binary CSP),  \textit{SPOT5} (satellite scheduling), \textit{driverlog} (planning in temporal and metric domains), \textit{grid} (Markov Random Field), normalized (MIPLib),  \textit{pedigree} (genetic Linkage) and \textit{CELAR} (frequency assignment).

\subsection{Results}

Our first analysis is about the impact of using or not using cost-function merging. Table~\ref{table:mergevsnonmerge} 
reports the relative time performance gain of doing cost-function merging. For each benchmark, speed-up is the solving time ratio of its best performing algorithm out of the 16 alternatives with and without cost-function merging. We observe that cost-function merging is consistently useful producing significant speed-ups that in some cases are over 250. The only case where cost-function merging is not advantageous is with \textit{Grid} instances where its use causes none of the 16 algorithms to solve any instance (not even with a larger time limit of 4 hours). Interestingly, without cost-function merging IHS-ub cores can solve all the instances with maximal and cost-bounded cores. The most probable reason is that the grid structure is so regular that the tree-decompostion used to decide which functions to merge is not appropriate.
Because this result is so conclusive, and for the sake of clarity, in the following every table reports results with cost-merging except for Grid where the reported results are without cost-merging.

\begin{table}[h]
\begin{center}
\begin{tabular}{ l|c } 
 \textbf{Problem} & \textbf{Speed-up} \\
 \hline
 Rnd domains &  268.31 \\
 Rnd weights &  2.18\\
 Rnd sparse & 39.33 \\
 Rnd scale-free-4 & 93.70 \\
 Rnd scale-free-5 & 12.69 \\
 Ehi &  99.53\\
 SPOT5 & 156.74 \\
 driverlog & 1.61 \\
 Grid &  0.46\\ 
 Normalized & 1.32 \\ 
 Pedigree &  1.34\\ 
 CELAR &   4.85\\ 
\end{tabular}
\end{center}
    \caption{Relative time performance gain of cost-function merging. On a given benchmark, speed-up is the solving time ratio of its best performing algorithm  with and without cost-function merging.       
    }
    \label{table:mergevsnonmerge}
\end{table}

 Tables~\ref{table:time-merge} and~\ref{table:comparisontablecore} report for each one of the 12 problem classes and each one of the 16 algorithms, the relative performance with respect to time and space, respectively. Each table entry is the ratio w.r.t. the best-performing algorithm on that benchmark. For example, in Table~\ref{table:time-merge}, a $1$ identifies the fastest algorithm while a value of $r$ indicates that the algorithm is $r$ times slower than the best.


Our next analysis is about the impact that each algorithm has on the solving time (Table~\ref{table:time-merge}).  Our first observation is that different algorithms have very different running times, but no algorithm dominates the others. In some benchmarks, that difference is so extreme that the best approach solves all instances while other approaches are not able to solve any of the instances within the time limit. 
If we look at problem classes where every instance is solved with all (or nearly all) algorithms and compare speed-up ratios, we still see that the best approach is at least $4$ times faster than the worst and the different can go up to several orders of magnitude. We also observe that some benchmarks are very sensitive with respect to the method in which hitting vectors are computed (e.g. \textit{Grid}, \textit{scale-free-5}), some benchmarks are very sensitive to the method in which cores are improved (e.g. \textit{CELAR}) and some benchmarks are very sensitive to both (e.g. \textit{SPOT5}).
Therefore, one should use caution before concluding from experiments that the HS approach is not suitable for a particular type of problem, because it may happen that the right algorithm has not been considered. 

Regarding the different ways to compute hitting vectors, we observe that the best option is again highly benchmark dependent. The best algorithm is 5 times with IHS-grdub, 3 times with IHS-ub and 4 times with IHS-lb. IHS-grdlb never appears in the best algorithm, but when the best algorithm is IHS-lb, it usually performs very closely. We observed that in the many cases in which greedy vectors are not beneficial the reason is that except for the very first iterations, greedy hitting vectors produce cheap but useless iterations and IHS-$grdlb$ (resp. IHS-$grdub$) converges to IHS-$lb$ (resp. IHS-$ub$). From that we conjecture that making more effective greedy algorithms, even at the cost of being more time consuming may be a useful improvement.

Regarding the different ways to improve cores, we observe that the most common best option is to compute maximal cores. The only two exceptions are \textit{Pedigree} and \textit{random-domains} where the best option is to compute lazy cores. The reason for the exception is that in these two benchmarks the induced CSPs are very difficult for the SAT solver and it pays off to generate much larger sets of cores even if it is at the cost of making the computation of hitting vectors more difficult. This observation let us believe that a promising improvement for this type of instances would be to find more efficient SAT solvers or even switching to some other solving paradigm.

It is surprising that the two extreme core improvement methods (maximal and lazy cores) are  best options and the intermediate methods never are so. Inspecting the results with more detail we observe that computing cost-bounded cores is often equivalent to not making any core improvement because the core that the assumption-based SAT solver provides for free already has a cost larger than the $ub$. We also observed that the cost of partially maximal cores is often too close to the cost of lazy cores (for example in \textit{normalized}), making them to weak.

Our third analysis is about the impact that each algorithm has on the number of cores $|\C|$ needed to achieve the Termination Condition $\GC \leq \C$  (Table~\ref{table:comparisontablecore}). The final size of $\C$ is relevant because it is related to both the number of iterations and the space requirements. 

Regarding hitting vector computation, the pattern is clear. As expected, IHS-$lb$ (optimal hitting vectors) requires fewer cores than IHS-$ub$ cores to dominate the set of goal cores, but the difference does not seem to be dramatic. 
The same thing happens when considering greedy hitting vectors.
From that, we conjecture that cost-bounded cores quickly become nearly as good as optimal cores (probably because the $ub$ gets tight) and the cost of greedy cores quickly becomes too high and therefore they become useless.

Regarding core improvement, the pattern is also clear. As expected, the weaker the improvement, the more cores are needed. Compared with maximal cores, the use of partially maximally maximal cores needs a larger $|\C|$ but the difference usually is not very large. 
When considering cost-bounded cores the set $\C$ ends up being much larger (e.g. \textit{Ehi}), which means that the effort of improving cores until their cost reaches the upper bound is not enough in general to dominate more goal cores and accelerate the termination of the execution. As a matter of fact, using cost-bounded cores has an effect very close to using lazy cores. A probable reason is that the cores provided by the assumption-based SAT solver often satisfy that their cost reaches the $ub$, so both approaches end up being equivalent.

From the experiments presented in this paper (and many others that we have also conducted), we can say that the effectiveness of IHS algorithms still falls far behind the state-of-the-art \textit{Toulbar2} solver~\cite{DBLP:journals/constraints/HurleyOAKSZG16}. However, in our experiments we see that there are instances where IHS is competitive. For instance, Table~\ref{table:ihsvstoulbar} compares average solving times and number of unsolved instances with Toulbar2 (version 1.2) and the best IHS among all the alternatives that we considered. We can see that IHS outperforms toulbar2 in Ehi, SPOT5, Grid, Normalized and Pedigree, but toulbar2 outperforms IHS in all the random benchmarks. All the instances where IHS seems suitable have in common small domains and not too many different costs in their cost functions. Small domains are good for solving induced CSPs encoding and solving them with SAT. Not too many different costs means smaller vector spaces. The conjunction of these two features seems a very reasonable proxy for IHS suitability and designing improvements for problems having this features may be the fastest route to find regions where IHS may be competitive.



\begin{landscape}
\begin{table}[h]
\centering
\begin{tabular}{l|rrrr|rrrr}
 & \bf{grdub} & \bf{ub} & \bf{grdlb} & \bf{lb} & \bf{grdub} & \bf{ub} & \bf{grdlb} & \bf{lb}\\ \hline
& \multicolumn{4}{c}{SPOT5 (13)} &  \multicolumn{4}{|c}{Pedigree (21)} \\
\hline

 \bf{lazy} & 25.50 (1) & 28.83 (1) & 41.73 (0) & 82.29 (2) & 1 (0) & 3.91 (0) & 3.72 (1) & 4.07 (2) \\ 
 \bf{cost-bounded} & 27.26 (1) & 29.19 (1) & 52.99 (0) & 80.58 (3) & 3.34 (0) & 3.63 (0) & 3.56 (1) & 3.71 (2) \\ 
 \bf{partially maximal} & 22.45 (1) & 22.31 (1) & 7.62 (0) & 7.62 (0) & 3.52 (0) & 3.62 (0) & 3.68 (0) & 3.58 (1) \\ 
 \bf{maximal} & 21.82 (1) & 21.74 (1) & 1.97 (0) & 1 (0) & 3.52 (0) & 3.37 (0) & 3.47 (0) & 3.26 (0) \\ 

\hline
& \multicolumn{4}{c}{driverlog (23)} &  \multicolumn{4}{|c}{CELAR (9)} \\
\hline  
 \bf{lazy} & 217.00 (6) & 298.69 (8) & 296.86 (6) & 408.44 (12) & 7.52 (4) & 11.29 (7) & 8.41 (5) & 11.29 (7) \\ 
 \bf{cost-bounded} & 210.87 (4) & 285.06 (7) & 293.31 (7) & 414.61 (13) & 7.68 (4) & 11.29 (7) & 8.46 (5) & 7.73 (4) \\ 
 \bf{partially maximal} & 56.86 (1) & 54.70 (1) & 62.74 (1) & 99.32 (1) & 1.86 (1) & 4.95 (3) & 1.82 (1) & 5.04 (3) \\ 
 \bf{maximal} & 1.25 (0) & 1.23 (0) & 1.36 (0) & 1 (0) & 1.78 (1) & 1.83 (1) & 1.72 (1) & 1 (0) \\ 

 \hline
& \multicolumn{4}{c}{Grid (5)} &  \multicolumn{4}{|c}{Ehi (200)} \\
\hline  
 \bf{lazy}  & 2.18 (5) & 1.30 (1) & 2.18 (5) & 2.18 (5) & 14.34 (0) & 21.65 (0) & 21.16 (0) & 114.02 (13) \\ 
 \bf{cost-bounded}  & 2.18 (5) & 1.42 (0) & 2.18 (5) & 2.18 (5) & 14.44 (0) & 21.40 (0) & 21.03 (0) & 113.56 (10) \\ 
 \bf{partially maximal} & 2.18 (5) & 1.05 (1) & 2.18 (5) & 2.18 (5) & 1.41 (0) & 1.28 (0) & 1.48 (0) & 1.51 (0) \\ 
 \bf{maximal} & 2.18 (5) & 1 (0) & 2.18 (5) & 2.18 (5) & 1.02 (0) & 1 (0) & 1.10 (0) & 1.04 (0) \\ 
   \hline
& \multicolumn{4}{c}{Normalized (4)} &  \multicolumn{4}{|c}{Rnd scale-free-4 (20)} \\
\hline  
 \bf{lazy} & 2.56 (2) & 2.82 (2) & 2.56 (2) & 3.82 (3) & 90.77 (18) & 91.97 (18) & 93.64 (19) & 98.50 (20) \\ 
 \bf{cost-bounded} & 2.56 (2) & 2.82 (2) & 2.56 (2) & 3.82 (3) & 90.23 (18) & 91.09 (18) & 93.63 (19) & 98.50 (20) \\ 
 \bf{partially maximal} & 2.55 (2) & 2.55 (2) & 2.55 (2) & 2.56 (2) & 37.04 (3) & 34.77 (3) & 81.75 (14) & 89.33 (17) \\ 
 \bf{maximal} & 1 (0) & 1.44 (0) & 2.32 (1) & 2.55 (2) & 1.24 (0) & 1 (0) & 16.39 (1) & 19.18 (1) \\ 

   \hline
& \multicolumn{4}{c}{Rnd scale-free-5  (20)} &  \multicolumn{4}{|c}{Rnd weights (50)} \\
\hline  
 \bf{lazy} & 12.69 (20) & 12.69 (20) & 12.69 (20) & 12.69 (20) & 4.09 (49) & 4.11 (50) & 4.01 (48) & 4.06 (49) \\ 
 \bf{cost-bounded} & 12.69 (20) & 12.69 (20) & 12.69 (20) & 12.69 (20) & 4.11 (50) & 4.11 (50) & 4.01 (47) & 4.06 (49) \\ 
 \bf{partially maximal} & 8.94 (11) & 8.66 (11) & 12.11 (19) & 12.23 (19) & 3.65 (41) & 3.68 (41) & 2.80 (27) & 2.84 (29) \\ 
 \bf{maximal} & 1 (0) & 1.01 (0) & 7.83 (9) & 7.68 (9) & 2.63 (21) & 3.17 (28) & 1.34 (4) & 1 (1) \\ 
 \hline
& \multicolumn{4}{c}{Rnd domains (50)} &  \multicolumn{4}{|c}{Rnd sparse (50)} \\
\hline  
 \bf{lazy} & 1 (0) & 3.28 (0) & 1.28 (0) & 4.35 (0) & 75.91 (19) & 97.88 (24) & 82.13 (20) & 143.50 (37) \\ 
 \bf{cost-bounded} & 1 (0) & 1.35 (0) & 1.11 (0) & 2.58 (0) & 73.87 (16) & 98.75 (23) & 80.05 (20) & 143.88 (36) \\ 
 \bf{partially maximal} & 3.35 (0) & 1.30 (0) & 3.41 (0) & 2.31 (0) & 22.89 (3) & 34.39 (5) & 40.11 (7) & 98.25 (24) \\ 
 \bf{maximal} & 3.80 (0) & 2.60 (0) & 3.95 (0) & 3.12 (0) & 1 (0) & 2.18 (0) & 6.08 (0) & 21.25 (2) \\ 

 \hline
\multicolumn{9}{c}{}\\
\end{tabular}
\caption{Relative time performance of 16 different IHS algorithms on 12 different problem classes using cost-function merging (except for the Grid class). For each class (in parenthesis its number of instances), columns indicate the method to compute hitting vectors and rows indicate the method to improve them. Each entry is the ratio of its average solving time over the average solving time of the best performing alternative in that benchmark (in parenthesis the number of unsolved instances within the time limit of $1$ hour).
}
\label{table:time-merge}
\end{table}
\end{landscape}

\begin{table}[t]
\centering
\begin{tabular}{l|rrrr|rrrr}
 & \bf{grdub} & \bf{ub} & \bf{grdlb} & \bf{lb} & \bf{grdub} & \bf{ub} & \bf{grdlb} & \bf{lb}\\ \hline
& \multicolumn{4}{c}{SPOT5} &  \multicolumn{4}{|c}{Pedigree} \\
\hline
\bf{lazy} & 6.23 & 6.33 & 5.94 & 6.74 & 1.55 & 1.36 & 1.50 & 1.27 \\ 
 \bf{cost-bounded} & 6.46 & 6.80 & 5.68 & 6.11 & 1.14 & 1.17 & 1.17 & 1.15 \\ 
 \bf{partially maximal} & 3.04 & 2.35 & 2.73 & 2.24 & 1.25 & 1.17 & 1.23 & 1.10 \\ 
 \bf{maximal} & 1.67 & 1.57 & 1.27 & 1 & 1.13 & 1.05 & 1.09 & 1 \\ 
\hline
& \multicolumn{4}{c}{driverlog} &  \multicolumn{4}{|c}{CELAR} \\
\hline  
 \bf{lazy} & 31.80 & 40.77 & 23.24 & 21.30 & 86.75 & 32.23 & 80.09 & 8.90 \\ 
 \bf{cost-bounded} & 29.96 & 39.08 & 22.46 & 21.90 &85.75 & 31.17 & 81.91 & 5.23 \\ 
 \bf{partially maximal} & 10.17 & 8.67 & 9.66 & 8.49 & 20.91 & 9.27 & 15.87 & 3.44 \\ 
 \bf{maximal} & 2.08 & 1.25 & 2.00 & 1 & 6.27 & 3.68 & 5.36 & 1 \\ 

 \hline
& \multicolumn{4}{c}{Grid} &  \multicolumn{4}{|c}{Ehi} \\
\hline  

 \bf{lazy} & 87.18 & 6.61 & 87.27 & 1.06 &  282.08 & 181.58 & 282.00 & 290.17 \\ 
 \bf{cost-bounded} & 69.33 & 6.61 & 86.72 & 1.05 &  282.53 & 181.98 & 281.83 & 290.82 \\ 
 \bf{partially maximal} & 86.93 & 7.88 & 86.90 & 1.04 &  7.05 & 4.95 & 7.03 & 5.84 \\ 
 \bf{maximal} & 84.41 & 7.61 & 84.55 & 1 &  1.16 & 1 & 1.19 & 1.10 \\ 
   \hline
   
& \multicolumn{4}{c}{Normalized} &  \multicolumn{4}{|c}{Rnd scale-free-4} \\
\hline  
 \bf{lazy} & 10.95 & 5.05 & 6.27 & 1.93 &  23.34 & 21.93 & 7.63 & 4.68 \\ 
 \bf{cost-bounded} & 11.41 & 5.03 & 5.66 & 1.93 & 24.21 & 22.19 & 7.76 & 4.58 \\ 
 \bf{partially maximal} & 6.28 & 3.64 & 3.62 & 1.16 & 7.91 & 7.55 & 3.81 & 3.04 \\ 
 \bf{maximal} & 2.43 & 1.49 & 1.97 & 1 & 1.48 & 1.24 & 1.29 & 1 \\ 
   \hline
& \multicolumn{4}{c}{Rnd scale-free-5} &  \multicolumn{4}{|c}{Rnd weights} \\
\hline  
 \bf{lazy} & 15.38 & 15.62 & 3.56 & 3.37 &  17.35 & 1.63 & 33.79 & 16.88 \\ 
 \bf{cost-bounded} & 15.16 & 15.86 & 3.35 & 3.32 & 19.91 & 1.69 & 37.05 & 16.61 \\ 
 \bf{partially maximal} & 9.68 & 9.49 & 2.11 & 1.95 &  4.29 & 1.47 & 10.89 & 8.38 \\ 
 \bf{maximal} & 2.07 & 2.06 & 1.18 & 1 & 2.28 & 1 & 2.58 & 1.38 \\ 
   \hline
& \multicolumn{4}{c}{Rnd domains} &  \multicolumn{4}{|c}{Rnd sparse } \\
\hline  
 \bf{lazy} & 15.29 & 8.87 & 15.84 & 9.43 & 95.52 & 15.14 & 89.26 & 4.22 \\ 
 \bf{cost-bounded} & 10.69 & 5.30 & 10.79 & 5.86 & 95.48 & 15.12 & 83.46 & 4.15 \\ 
 \bf{partially maximal} & 8.02 & 2.28 & 7.80 & 3.38 & 20.97 & 6.60 & 21.50 & 2.81 \\ 
 \bf{maximal} & 1.55 & 1 & 1.58 & 1.07 & 2.67 & 1.69 & 2.56 & 1 \\ 
\hline
\multicolumn{9}{c}{}\\
\end{tabular}
\caption{Relative space performance (measured as the size of the set of cores at termination or at time out) of 16 different IHS algorithms on 12 different problem classes using cost-function merging. Each entry is the ratio average $|\C|$ divided by the minimum $|\C|$ among the algorithms.}\label{table:comparisontablecore}
\end{table}

\begin{table}[h]
\begin{center}
\begin{tabular}{ l|cc|cc } 
 \multirow{2}{*}{\textbf{Problem}} & \multicolumn{2}{c|}{\textbf{Toulbar2}} & \multicolumn{2}{c}{\textbf{IHS (best)}} \\
 \cline{2-5}
 
  & \bf{time} & \bf{nb. unsolved} &  \bf{time} & \bf{nb. unsolved}\\
 \hline
 Rnd domains &  0.65 & 0 & 13.42 & 0 \\
 Rnd weights &  0.01 & 0 & 509.04 & 1 \\
 Rnd sparse & 0.02 & 0 & 20.15 & 0 \\
 Rnd scale-free-4 & 0.01 & 0 & 36.55 & 0 \\
 Rnd scale-free-5 & 0.03 & 0 & 283.73 & 0 \\
 Ehi &  136.57 & 1 & 10.86 & 0 \\
 SPOT5 & 2367.06 & 8 & 13.25 & 0 \\
 driverlog & 0.45 & 0 & 3.80 & 0 \\
 Grid & 3600 & 5 & 1652.77 & 0 \\ 
 Normalized & 912.42 & 1 & 707.93 & 0 \\ 
 Pedigree &  631.39 & 3 & 81.96 & 0 \\ 
 CELAR &  0.46 & 0 & 248.06 & 0 \\ 
\end{tabular}
\end{center}
    \caption{Average running times (in seconds) and number of unsolved instances within the time limit of 1 hour. Toulbar2 is executed with default parameters. For each benchmark, IHS is its best among all the alternatives. 
    }
    \label{table:ihsvstoulbar}
\end{table}

\section{Conclusion and Future Work}
We have presented a large empirical evaluation of 32 alternative implementations of the Implicit Hitting Set approach for WCSP solving.
Although our current implementations of IHS are only competitive with the state-of-the-art Toulbar2 solver in selected instances, our results show how different is the performance of different alternatives, and we believe that it indicates
that the approach is very general and has potential.

We covered a variety of alternatives, but many known improvements that have been found useful in other paradigms remain to be adapted to the WCSP framework and tested. For example, we want to consider in the future reduced cost fixing or weight-aware cost extraction~\cite{DBLP:conf/sat/IHS-PB2}.
It is reasonable to expect that the IHS will also benefit from them in the WCSP paradigm.

More importantly, we believe that all the components of the algorithms that we have tested can be improved. We
want to evaluate alternative ways to solve induced CSPs and a natural option is to replace the SAT solver by a constraint programming solver. We also want to evaluate alternative ways to find cost-bounded hitting vectors and a natural option would be to replace CPLEX by a Pseudo-Boolean optimization solver or a SAT solver with one of the many efficient encodings of Pseudo-Boolean constraints. Finally, we want to evaluate alternatives to the greedy algorithm and we believe that local search is a promising direction.




\bibliography{biblio}

\begin{thebibliography}{10}

\bibitem{DBLP:conf/cp/AlloucheGKSZ15}
David Allouche, Simon de~Givry, George Katsirelos, Thomas Schiex, and Matthias Zytnicki.
\newblock Anytime hybrid best-first search with tree decomposition for weighted {CSP}.
\newblock In Gilles Pesant, editor, {\em Principles and Practice of Constraint Programming - 21st International Conference, {CP} 2015, Cork, Ireland, August 31 - September 4, 2015, Proceedings}, volume 9255 of {\em Lecture Notes in Computer Science}, pages 12--29. Springer, 2015.
\newblock \href {https://doi.org/10.1007/978-3-319-23219-5\_2} {\path{doi:10.1007/978-3-319-23219-5\_2}}.

\bibitem{DBLP:journals/algorithms/AnsoteguiBL22}
Carlos Ans{\'{o}}tegui, Maria~Luisa Bonet, and Jordi Levy.
\newblock Scale-free random {SAT} instances.
\newblock {\em Algorithms}, 15(6):219, 2022.
\newblock URL: \url{https://doi.org/10.3390/a15060219}, \href {https://doi.org/10.3390/A15060219} {\path{doi:10.3390/A15060219}}.

\bibitem{DBLP:conf/cp/BeldjilaliMAKG22}
Abdelkader Beldjilali, Pierre Montalbano, David Allouche, George Katsirelos, and Simon de~Givry.
\newblock Parallel hybrid best-first search.
\newblock In Christine Solnon, editor, {\em 28th International Conference on Principles and Practice of Constraint Programming, {CP} 2022, July 31 to August 8, 2022, Haifa, Israel}, volume 235 of {\em LIPIcs}, pages 7:1--7:10. Schloss Dagstuhl - Leibniz-Zentrum f{\"{u}}r Informatik, 2022.
\newblock URL: \url{https://doi.org/10.4230/LIPIcs.CP.2022.7}, \href {https://doi.org/10.4230/LIPICS.CP.2022.7} {\path{doi:10.4230/LIPICS.CP.2022.7}}.

\bibitem{DBLP:journals/constraints/BensanaLV99}
E.~Bensana, Michel Lema{\^{\i}}tre, and G{\'{e}}rard Verfaillie.
\newblock Earth observation satellite management.
\newblock {\em Constraints An Int. J.}, 4(3):293--299, 1999.
\newblock \href {https://doi.org/10.1023/A:1026488509554} {\path{doi:10.1023/A:1026488509554}}.

\bibitem{DBLP:conf/sat/BergBP20}
Jeremias Berg, Fahiem Bacchus, and Alex Poole.
\newblock Abstract cores in implicit hitting set maxsat solving.
\newblock In Luca Pulina and Martina Seidl, editors, {\em Theory and Applications of Satisfiability Testing - {SAT} 2020 - 23rd International Conference, Alghero, Italy, July 3-10, 2020, Proceedings}, volume 12178 of {\em Lecture Notes in Computer Science}, pages 277--294. Springer, 2020.
\newblock \href {https://doi.org/10.1007/978-3-030-51825-7\_20} {\path{doi:10.1007/978-3-030-51825-7\_20}}.

\bibitem{cadical}
Armin Biere, Katalin Fazekas, Mathias Fleury, and Maximillian Heisinger.
\newblock {CaDiCaL}, {Kissat}, {Paracooba}, {Plingeling} and {Treengeling} entering the {SAT Competition 2020}.
\newblock In Tomas Balyo, Nils Froleyks, Marijn Heule, Markus Iser, Matti J{\"a}rvisalo, and Martin Suda, editors, {\em Proc.~of {SAT Competition} 2020 -- Solver and Benchmark Descriptions}, volume B-2020-1 of {\em Department of Computer Science Report Series B}, pages 51--53. University of Helsinki, 2020.

\bibitem{DBLP:journals/constraints/CabonGLSW99}
Bertrand Cabon, Simon de~Givry, Lionel Lobjois, Thomas Schiex, and Joost~P. Warners.
\newblock Radio link frequency assignment.
\newblock {\em Constraints An Int. J.}, 4(1):79--89, 1999.
\newblock \href {https://doi.org/10.1023/A:1009812409930} {\path{doi:10.1023/A:1009812409930}}.

\bibitem{DBLP:journals/ai/CooperGSSZW10}
Martin~C. Cooper, Simon de~Givry, Mart{\'{\i}} S{\'{a}}nchez{-}Fibla, Thomas Schiex, Matthias Zytnicki, and Tom{\'{a}}s Werner.
\newblock Soft arc consistency revisited.
\newblock {\em Artif. Intell.}, 174(7-8):449--478, 2010.
\newblock \href {https://doi.org/10.1016/j.artint.2010.02.001} {\path{doi:10.1016/j.artint.2010.02.001}}.

\bibitem{cormen01introduction}
Thomas~H. Cormen, Charles~E. Leiserson, Ronald~L. Rivest, and Clifford Stein.
\newblock {\em Introduction to Algorithms}.
\newblock The MIT Press, 2nd edition, 2001.
\newblock URL: \url{http://www.amazon.com/Introduction-Algorithms-Thomas-H-Cormen/dp/0262032937%3FSubscriptionId%3D13CT5CVB80YFWJEPWS02%26tag%3Dws%26linkCode%3Dxm2%26camp%3D2025%26creative%3D165953%26creativeASIN%3D0262032937}.

\bibitem{cplex2009v12}
IBM~ILOG Cplex.
\newblock V12. 1: User’s manual for cplex.
\newblock {\em International Business Machines Corporation}, 46(53):157, 2009.

\bibitem{DBLP:phd/ca/Davies14}
Jessica Davies.
\newblock {\em Solving {MAXSAT} by Decoupling Optimization and Satisfaction}.
\newblock PhD thesis, University of Toronto, Canada, 2014.
\newblock URL: \url{http://hdl.handle.net/1807/43539}.

\bibitem{DBLP:conf/cp/DaviesB13}
Jessica Davies and Fahiem Bacchus.
\newblock Postponing optimization to speed up {MAXSAT} solving.
\newblock In Christian Schulte, editor, {\em Principles and Practice of Constraint Programming - 19th International Conference, {CP} 2013, Uppsala, Sweden, September 16-20, 2013. Proceedings}, volume 8124 of {\em Lecture Notes in Computer Science}, pages 247--262. Springer, 2013.
\newblock \href {https://doi.org/10.1007/978-3-642-40627-0\_21} {\path{doi:10.1007/978-3-642-40627-0\_21}}.

\bibitem{DBLP:conf/cp/DelisleB13}
Erin Delisle and Fahiem Bacchus.
\newblock Solving weighted csps by successive relaxations.
\newblock In Christian Schulte, editor, {\em Principles and Practice of Constraint Programming - 19th International Conference, {CP} 2013, Uppsala, Sweden, September 16-20, 2013. Proceedings}, volume 8124 of {\em Lecture Notes in Computer Science}, pages 273--281. Springer, 2013.
\newblock \href {https://doi.org/10.1007/978-3-642-40627-0\_23} {\path{doi:10.1007/978-3-642-40627-0\_23}}.

\bibitem{DBLP:conf/cade/FazekasBB18}
Katalin Fazekas, Fahiem Bacchus, and Armin Biere.
\newblock Implicit hitting set algorithms for maximum satisfiability modulo theories.
\newblock In Didier Galmiche, Stephan Schulz, and Roberto Sebastiani, editors, {\em Automated Reasoning - 9th International Joint Conference, {IJCAR} 2018, Held as Part of the Federated Logic Conference, FloC 2018, Oxford, UK, July 14-17, 2018, Proceedings}, volume 10900 of {\em Lecture Notes in Computer Science}, pages 134--151. Springer, 2018.
\newblock \href {https://doi.org/10.1007/978-3-319-94205-6\_10} {\path{doi:10.1007/978-3-319-94205-6\_10}}.

\bibitem{DBLP:books/fm/GareyJ79}
M.~R. Garey and David~S. Johnson.
\newblock {\em Computers and Intractability: {A} Guide to the Theory of NP-Completeness}.
\newblock W. H. Freeman, 1979.

\bibitem{DBLP:reference/fai/GentPP06}
Ian~P. Gent, Karen~E. Petrie, and Jean{-}Fran{\c{c}}ois Puget.
\newblock Symmetry in constraint programming.
\newblock In Francesca Rossi, Peter van Beek, and Toby Walsh, editors, {\em Handbook of Constraint Programming}, volume~2 of {\em Foundations of Artificial Intelligence}, pages 329--376. Elsevier, 2006.
\newblock \href {https://doi.org/10.1016/S1574-6526(06)80014-3} {\path{doi:10.1016/S1574-6526(06)80014-3}}.

\bibitem{DBLP:journals/constraints/HurleyOAKSZG16}
Barry Hurley, Barry O'Sullivan, David Allouche, George Katsirelos, Thomas Schiex, Matthias Zytnicki, and Simon de~Givry.
\newblock Multi-language evaluation of exact solvers in graphical model discrete optimization.
\newblock {\em Constraints An Int. J.}, 21(3):413--434, 2016.
\newblock \href {https://doi.org/10.1007/s10601-016-9245-y} {\path{doi:10.1007/s10601-016-9245-y}}.

\bibitem{Larrosa24}
Javier Larrosa, Conrado Martinez, and Emma Rollon.
\newblock Theoretical and empirical analysis of cost-function merging for implicit hitting set wcsp solving.
\newblock In {\em The 38th Annual AAAI Conference on Artificial Intelligence February 20-27, 2024; Vancouver, Canada}. AAAI Press, 2024.

\bibitem{DBLP:journals/ai/LarrosaS04}
Javier Larrosa and Thomas Schiex.
\newblock Solving weighted {CSP} by maintaining arc consistency.
\newblock {\em Artif. Intell.}, 159(1-2):1--26, 2004.
\newblock \href {https://doi.org/10.1016/j.artint.2004.05.004} {\path{doi:10.1016/j.artint.2004.05.004}}.

\bibitem{DBLP:conf/sat/MUSalg}
Jo{\~{a}}o Marques{-}Silva and In{\^{e}}s Lynce.
\newblock On improving {MUS} extraction algorithms.
\newblock In Karem~A. Sakallah and Laurent Simon, editors, {\em Theory and Applications of Satisfiability Testing - {SAT} 2011 - 14th International Conference, {SAT} 2011, Ann Arbor, MI, USA, June 19-22, 2011. Proceedings}, volume 6695 of {\em Lecture Notes in Computer Science}, pages 159--173. Springer, 2011.
\newblock \href {https://doi.org/10.1007/978-3-642-21581-0\_14} {\path{doi:10.1007/978-3-642-21581-0\_14}}.

\bibitem{DBLP:conf/kr/SaikkoDAJ18}
Paul Saikko, Carmine Dodaro, Mario Alviano, and Matti J{\"{a}}rvisalo.
\newblock A hybrid approach to optimization in answer set programming.
\newblock In Michael Thielscher, Francesca Toni, and Frank Wolter, editors, {\em Principles of Knowledge Representation and Reasoning: Proceedings of the Sixteenth International Conference, {KR} 2018, Tempe, Arizona, 30 October - 2 November 2018}, pages 32--41. {AAAI} Press, 2018.
\newblock URL: \url{https://aaai.org/ocs/index.php/KR/KR18/paper/view/18021}.

\bibitem{DBLP:conf/cp/IHS-PB1}
Pavel Smirnov, Jeremias Berg, and Matti J{\"{a}}rvisalo.
\newblock Pseudo-boolean optimization by implicit hitting sets.
\newblock In Laurent~D. Michel, editor, {\em 27th International Conference on Principles and Practice of Constraint Programming, {CP} 2021, Montpellier, France (Virtual Conference), October 25-29, 2021}, volume 210 of {\em LIPIcs}, pages 51:1--51:20. Schloss Dagstuhl - Leibniz-Zentrum f{\"{u}}r Informatik, 2021.
\newblock URL: \url{https://doi.org/10.4230/LIPIcs.CP.2021.51}, \href {https://doi.org/10.4230/LIPICS.CP.2021.51} {\path{doi:10.4230/LIPICS.CP.2021.51}}.

\bibitem{DBLP:conf/sat/IHS-PB2}
Pavel Smirnov, Jeremias Berg, and Matti J{\"{a}}rvisalo.
\newblock Improvements to the implicit hitting set approach to pseudo-boolean optimization.
\newblock In Kuldeep~S. Meel and Ofer Strichman, editors, {\em 25th International Conference on Theory and Applications of Satisfiability Testing, {SAT} 2022, August 2-5, 2022, Haifa, Israel}, volume 236 of {\em LIPIcs}, pages 13:1--13:18. Schloss Dagstuhl - Leibniz-Zentrum f{\"{u}}r Informatik, 2022.
\newblock URL: \url{https://doi.org/10.4230/LIPIcs.SAT.2022.13}, \href {https://doi.org/10.4230/LIPICS.SAT.2022.13} {\path{doi:10.4230/LIPICS.SAT.2022.13}}.

\bibitem{DBLP:journals/bioinformatics/ViricelGSB18}
Cl{\'{e}}ment Viricel, Simon de~Givry, Thomas Schiex, and Sophie Barbe.
\newblock Cost function network-based design of protein-protein interactions: predicting changes in binding affinity.
\newblock {\em Bioinform.}, 34(15):2581--2589, 2018.
\newblock \href {https://doi.org/10.1093/bioinformatics/bty092} {\path{doi:10.1093/bioinformatics/bty092}}.

\bibitem{DBLP:journals/bioinformatics/VucinicSRBS20}
Jelena Vucinic, David Simoncini, Manon Ruffini, Sophie Barbe, and Thomas Schiex.
\newblock Positive multistate protein design.
\newblock {\em Bioinform.}, 36(1):122--130, 2020.
\newblock URL: \url{https://doi.org/10.1093/bioinformatics/btz497}, \href {https://doi.org/10.1093/BIOINFORMATICS/BTZ497} {\path{doi:10.1093/BIOINFORMATICS/BTZ497}}.

\end{thebibliography}

\end{document}